\documentclass[runningheads,table,xcdraw]{llncs}
\usepackage{graphicx}

\usepackage{tikz}
\usepackage{comment}
\usepackage{amsmath,amssymb} 
\usepackage{color}

\usepackage{subcaption} 
\usepackage{microtype}
\usepackage{cite}
\usepackage{amsmath,amssymb}
\usepackage{amsfonts}

\usepackage{enumitem} 

\usepackage[pagebackref,breaklinks,colorlinks]{hyperref} 

\usepackage{booktabs}
\usepackage{graphicx}

\usepackage[accsupp]{axessibility}  

\usepackage[width=122mm,left=12mm,paperwidth=146mm,height=193mm,top=12mm,paperheight=217mm]{geometry}

\usepackage{microtype}
\usepackage{cite}
\usepackage{amsmath,amssymb}
\usepackage{amsfonts}

\usepackage{enumitem} 

\usepackage[pagebackref,breaklinks,colorlinks]{hyperref} 

\usepackage{booktabs}
\usepackage{graphicx}

\usepackage{enumitem}
\usepackage{soul} 

\usepackage{pdfrender}
\newcommand*{\boldcheckmark}{%
  \textpdfrender{
    TextRenderingMode=FillStroke,
    LineWidth=.5pt, 
  }{\checkmark}%
}

\usepackage{algorithm}
\usepackage{algpseudocode}
\algrenewcommand\algorithmicrequire{\textbf{Input:}}
\algrenewcommand\algorithmicensure{\textbf{Output:}}
\usepackage{tabularx}
\makeatletter
\newcommand{\multiline}[1]{%
  \begin{tabularx}{\dimexpr\linewidth-\ALG@thistlm}[t]{@{}X@{}}
    #1
  \end{tabularx}
}
\makeatother

\def\onedot{\ifx\@let@token.\else.\null\fi\xspace}

\makeatother

\begin{document}
\pagestyle{headings}
\mainmatter
\def\ECCVSubNumber{10}  

\title{Seeing Objects in dark with Continual Contrastive Learning\\}

\titlerunning{Seeing Objects in dark with Continual Contrastive Learning}
%
\author{
Ujjal Kr Dutta \orcidID{0000-0002-0470-5521}\index{Dutta, Ujjal Kr}
}
\authorrunning{Dutta et al.}
%
\institute{
Myntra, Bengaluru, India\\
\email{ukdacad@gmail.com}\\
}
\maketitle

\begin{abstract}
Object Detection, a fundamental computer vision problem, has paramount importance in smart camera systems. However, a truly reliable camera system could be achieved if and only if the underlying object detection component is robust enough across varying imaging conditions (or domains), for instance, different times of the day, adverse weather conditions, etc. In an effort to achieving a reliable camera system, in this paper, we make an attempt to train such a robust detector. Unfortunately, to build a well-performing detector across varying imaging conditions, one would require labeled training images (often in large numbers) from a plethora of corner cases. As manually obtaining such a large labeled dataset may be infeasible, we suggest using synthetic images, to mimic different training image domains. We propose a novel, contrastive learning method to align the latent representations of a pair of real and synthetic images, to make the detector robust to the different domains. However, we found that merely contrasting the embeddings may lead to catastrophic forgetting of the information essential for object detection. Hence, we employ a continual learning based penalty, to alleviate the issue of forgetting, while contrasting the representations. We showcase that our proposed method outperforms a wide range of alternatives to address the extremely challenging, yet under-studied scenario of object detection at night-time.%
\end{abstract}

\section{Introduction}
Smart camera systems are an integral part of crucial real-world applications such as surveillance systems, autonomous driving, medical imaging, to name a few. Object Detection, a fundamental computer vision problem, has paramount importance in smart camera systems. A reliable camera system depends on the robustness of the underlying object detection component, i.e., in the ability of the latter to detect objects across varying imaging conditions, or \textit{domains}. Let us say, we have deployed an object detection model which is trained to detect objects in daytime images with high mean Average Precision (mAP). However, even state-of-the-art object detection models perform poorly when an inference image belongs to a different domain (say, for example, night-time). This could greatly hinder the reliability of the overall camera system.

A potential workaround would be to collect a large pool of labeled data from different training scenarios or domains and retrain the model, which, at times, is infeasible. In this paper, to address the above problem, we propose a novel, Contrastive Learning approach to train a robust object detector. The key idea in Contrastive Learning is to align the latent representations of a pair of input images, which are semantically related. In our case, we suggest aligning the representations of a pair of real and synthetic images, of the same scene. The synthetic image is generated to mimic a different time of the day, corresponding to the scene of the original input image. Aligning the representations of the different times of the day helps in teaching the detector to learn those information which are essential to recognizing the objects, irrespective of the domain of an inference image.

Our method could be intuitively looked at as an implicit way of performing Domain Generalization (DG). While DG, as popularly studied in the recent classification literature, requires the presence of a number of \textbf{labeled} training domains to train a robust model (for evaluating on an unseen inference domain), our method makes no such assumption. Our method simply mimics the presence of different training domains by virtue of synthetic images. Without loss of generality, the synthetic images could be generated using any off-the-shelf method. However, we propose using a crafty, yet simple, Fourier transformation based method, which does not require any training!

While Unsupervised Domain Adaptation (UDA) techniques for object detection do leverage unlabeled data from the inference \textit{target} domain, they still require a large pool of unlabeled data, from the exact domain, from which the inference images would arrive from. Our method, on the other hand, is not dependent on the number of unlabeled images as well. This is because, with the Fourier transformation method, we could control a hyperparameter, and generate multiple synthetic images using a single seed image, randomly.

However, we found that merely contrasting the embeddings may lead to \textit{catastrophic forgetting} of the information essential for object detection. Hence, we employ a Continual Learning based penalty, to alleviate the issue of forgetting, while contrasting the representations. We showcase the effectiveness of our proposed method by evaluating it against a wide range of alternatives, to address the extremely challenging, yet under-studied scenario of object detection at night-time.

\section{Proposed Method}

\begin{figure*}
\centering
\begin{minipage}{.7\textwidth}
  \centering
	\includegraphics[width=\linewidth]{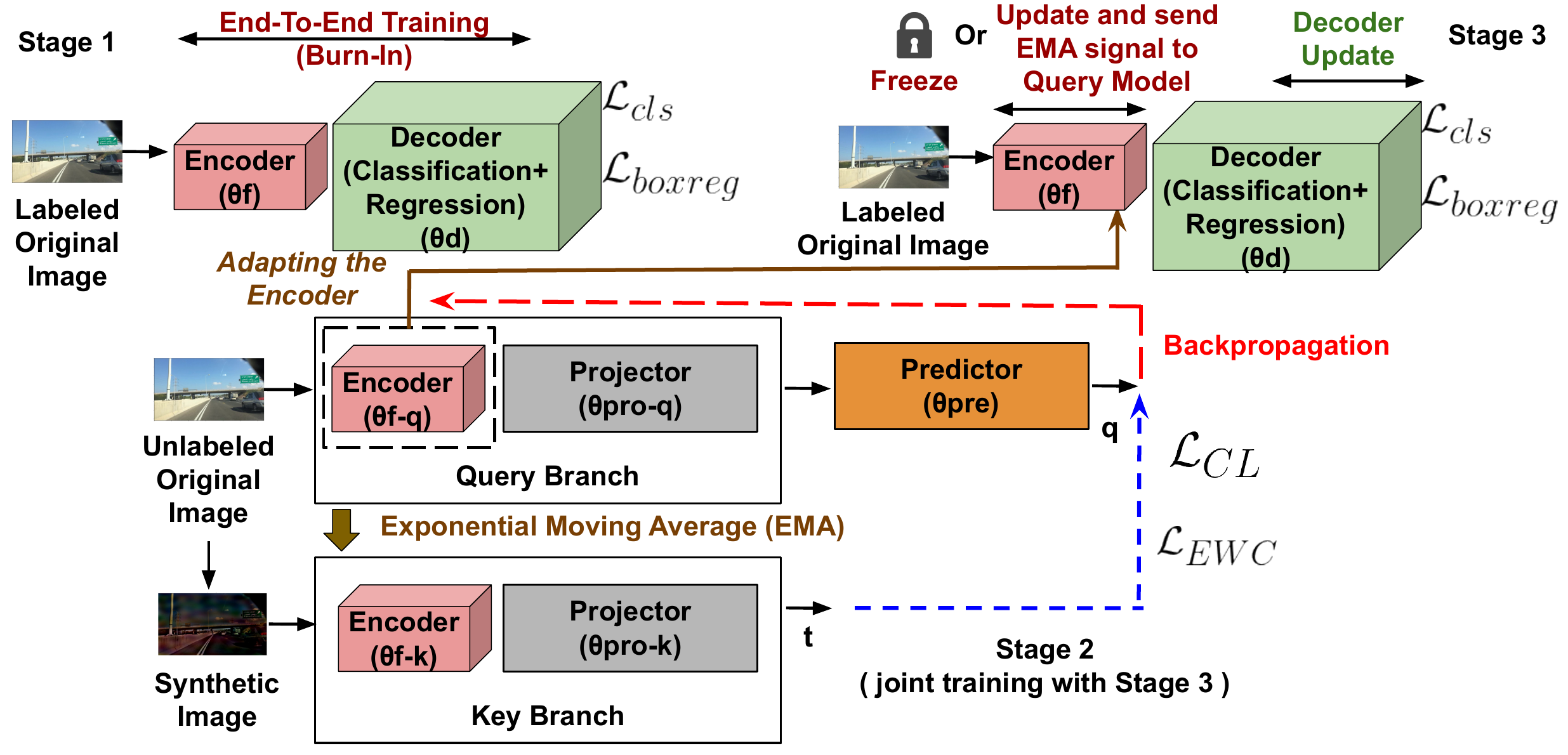}
    \caption{An illustration of the proposed approach. The input images belong to the BDD100K dataset. The figure is best viewed in color, when zoomed in.}
    \label{FCLUDA_OD}
\end{minipage}%
\hspace{0.1cm}
\begin{minipage}{.28\textwidth}
  \centering
  \includegraphics[width=\linewidth]{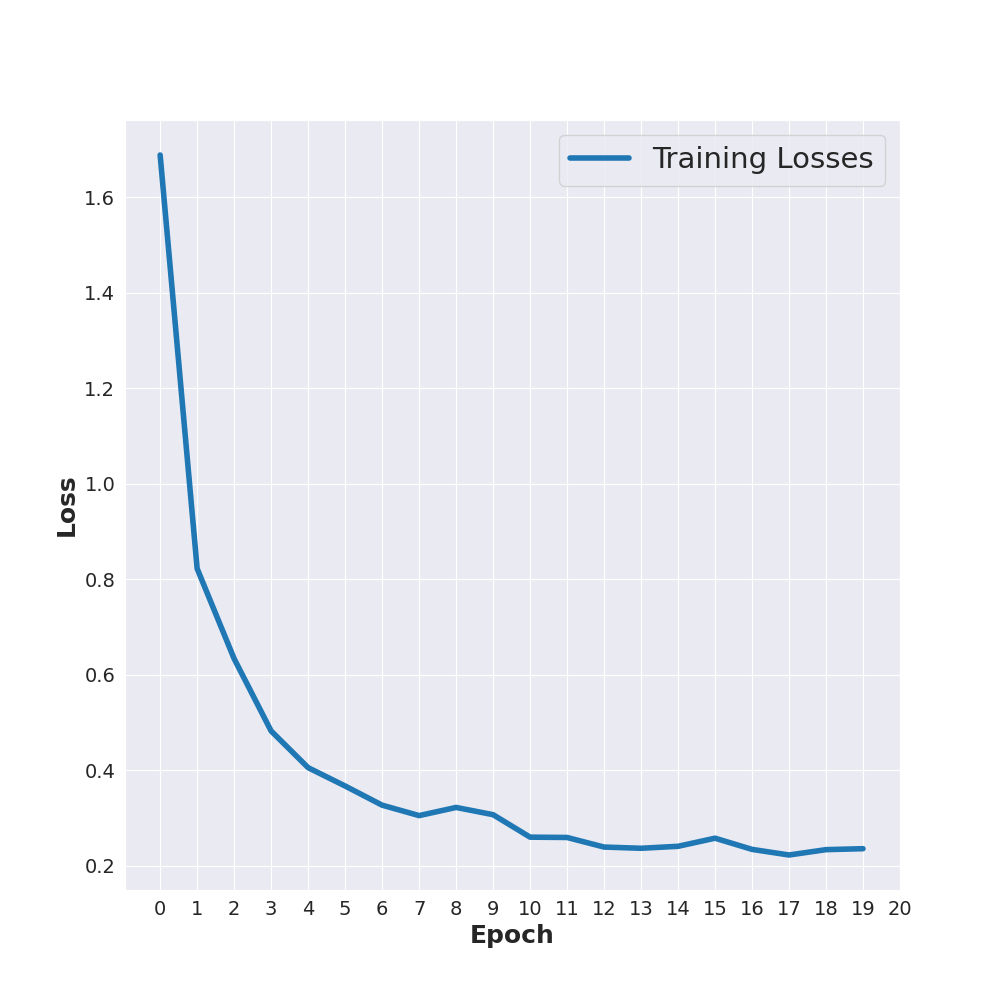}
  \captionof{figure}{Convergence behaviour of the contrastive loss with EWC.}
  \label{losses_cl}
\end{minipage}
\vspace{-0.8cm}
\end{figure*}

\textbf{Problem Description: } Let us assume that we are given a set of images $\{ x_i^d \}_{i=1}^{N_d}$, obtained from a source domain $\mathcal{D}_d$ (say, daytime), such that it is also feasible to manually annotate all the images with bounding boxes depicting the locations of the contained objects in them. Without loss of generality, we assume the presence of a deep object detection model whose layers are subdivided into an Encoder (to obtain a latent representation of a raw image), and a Decoder (consisting of classification and regression layers). Let, $\theta_f$ denote the parameters of the Encoder / Feature extractor, and $\theta_d$ refers to the parameters of the Decoder. We seek to learn $\theta_f$ and $\theta_d$ in such a way that the detector is robust across input images from a target domain, different from the source domain. Our proposed method is illustrated in Figure \ref{FCLUDA_OD}, and consists of three different stages.

\textbf{Stage 1 (Supervised Burn-in): } At first, we train our object detector (with parameters $\theta_f$ and $\theta_d$) end-to-end using the available annotated set $\{ x_i^d \}_{i=1}^{N_d}$ of images from $\mathcal{D}_d$. We refer to this as the \textit{burn-in} stage (denoted as Stage 1 in Figure \ref{FCLUDA_OD}). This stage involves optimizing a classification loss $\mathcal{L}_{cls}$ and a regression loss $\mathcal{L}_{boxreg}$.

\textbf{Stage 2-a (Unsupervised, Contrastive Learning): } In this stage, we try to adapt our encoder to be able to learn robust representations of raw images from across domains. For this, we want to ensure that the encoder focuses on learning information which is essential to recognize objects irrespective of the time of the day (i.e., domain). Hence, we generate synthetic images (details discussed in the experiments) to mimic different times of the day. Now, given a pair consisting of an original image and a synthetic image, we perform Contrastive Learning to align their representations, and enforce generalizability of the encoder towards images of different domains.

Given a raw image $x_i$, let $f_{\theta_f}(x_i)$ denotes the encoding obtained by the Encoder. For an arbitrary image $x_i^d \in \mathcal{D}_d$ we have its corresponding synthetic representation $x_i^{d\rightarrow s}$. The detector would be robust across domains if the underlying Encoder is able to learn similar representations for $f_{\theta_f}(x_i^d)$ and $f_{\theta_f}(x_i^{d\rightarrow s})$.

To learn similar representations for $f_{\theta_f}(x_i^d)$ and $f_{\theta_f}(x_i^{d\rightarrow s})$, we employ a Contrastive Learning module (Stage 2 of Figure \ref{FCLUDA_OD}). For this, we maintain a separate Siamese network having two branches, query and key. We mention the following key blocks essential for our formulation (with names of corresponding parameters within bracket):
\setlist{nolistsep}
\begin{enumerate}[noitemsep]
    \item \textbf{Query feature extractor} ($\theta_{f-q}$), consisting of the Encoder: It obtains an intermediate encoding of a raw image.
    \item \textbf{Query projector} ($\theta_{pro-q}$): Flattens the intermediate encoding to a vector embedding.
    \item \textbf{Key feature extractor} ($\theta_{f-k}$): Key branch's counterpart similar to Query feature extractor.
    \item \textbf{Key projector} ($\theta_{pro-k}$): Key branch's counterpart similar to Query projector.
\end{enumerate}
In order to make the overall detector robust across different times of the day, we aim at making $\theta_{f-q}$ robust across domains, and then later update $\theta_{f}$ using $\theta_{f-q}$. To do so, it suffices to maximize the inner-product similarity of the embeddings (representations) obtained after the Query and Key projectors, assuming that we pass $x_i^d \in \mathcal{D}_d$ (referring it as \textit{anchor}) and $x_i^{d\rightarrow s}$ (referring it as \textit{positive}) through the query and key branches respectively.

However, naively maximizing the inner-product similarity may lead to a trivial solution with a model collapse. Hence, to enforce asymmetry, following Self-Supervised Learning (SSL) \cite{BYOL_neurips20}, we add an additional predictor block with parameters $\theta_{pre}$ after the query, which obtains the \textbf{query embedding} $\mathbf{q}$. We call the embedding obtained by the key branch as the \textbf{target embedding} $\mathbf{t}$, as it provides a target/ guidance for $\mathbf{q}$. While $\theta_{f-q}$ and $\theta_{f-k}$ are initialized using the learned $\theta_f$ from Stage 1, the others, $\theta_{pro-q}$, $\theta_{pro-k}$ and $\theta_{pre}$ are initialized randomly. Let $\phi_q=(\theta_{f-q},\theta_{pro-q})$ denote parameters of the query branch, which are updated via backpropagating the gradients. The parameters of the key branch $\phi_k=(\theta_{f-k},\theta_{pro-k})$ are obtained using an Exponential Moving Average (EMA) \cite{meanteacher_neurips17}, as:
\begin{equation}
    \label{ema_FCL}
    \phi_k=\mu\phi_k + (1-\mu)\phi_q, \mu\in[0,1].
\end{equation}
The EMA (instead of backpropagation via the key branch) ensures stable updates of the key, which is essential to provide target to the query branch. Note that the Stage 2 is completely unsupervised in nature, meaning that it does not require manual bounding box annotations.  This stage involves optimizing a Contrastive Learning loss $\mathcal{L}_{CL}$.

\textbf{Stage 2-b Continual Learning with Elastic Weight Consolidation (EWC): }  However, naively optimizing a Contrastive Loss $\mathcal{L}_{CL}$ as above, although bridges the domain gap, leads to a catastrophic forgetting of the information learned during the burn-in stage. Thus, we add a penalty term $\mathcal{L}_{EWC}$ to $\mathcal{L}_{CL}$ that prohibits significant updates to the parameters important for the Task 1 (object detection/ burn-in), while learning Task 2 (Contrastive Learning in Stage 2). This penalty is based on Elastic Weight Consolidation (EWC) \cite{EWC_2017} enforced on the parameters of the query model. Using this penalty helps in continually learning domain invariant features while preserving the information learned apriori during burn-in.

Essentially, Task 1 in our case is to learn to ``detect objects" (in say, daytime, where enough labeled data is available). Task 2 does alignment of two domains (day and synthetic) by Contrastive Learning. But if we focus extensively on Task 2, we may deviate from the learned capabilities in Task 1 (this is catastrophic forgetting). To avoid this, we need continual learning, which incrementally learns Task 2, while maintaining consistency with Task 1.

\textbf{Stage 3 (Decoder Update): } Finally, in Stage 3, we replace the parameters $\theta_f$ of the Encoder, with that of the parameters $\theta_{f-q}$ of the query model, and update $\theta_d$ using the labeled images $\{ x_i^d \}_{i=1}^{N_d}$.

\textbf{Objective: } Our overall objective can thus be expressed as:
\begin{equation}
\label{overall_objective}
\begin{split}
    \min_{\theta_f,\theta_d,\theta_{f-q},\theta_{pro-q},\theta_{pre}} &(1-\lambda_{ewc})\mathcal{L}_{CL}+\lambda_{ewc}\mathcal{L}_{EWC}+\\
    &\mathcal{L}_{cls}+\mathcal{L}_{boxreg}.    
\end{split}
\end{equation}
Here, $\lambda_{ewc} \in (0,1)$ is a hyperparameter, $\mathcal{L}_{CL}=2-2\frac{\mathbf{q}^\top\mathbf{t}}{\left \| \mathbf{q} \right \|_2 \left \| \mathbf{t} \right \|_2}$ is the Contrastive Loss, $\mathcal{L}_{cls}$ and $\mathcal{L}_{boxreg}$ are the classification and bounding box regression losses of the base object detector. Also, in our case, we define the EWC based penalty as:
\begin{equation}
    \label{ewc_penalty}
    \mathcal{L}_{EWC}=\sum_p F_p({(\theta_{f-q})}_p-{(\theta_{f,Burn-in})}^*_p)^2.
\end{equation}
Here, $F$ is the Fisher Information matrix, ${(\theta_{f-q})}_p$ indicates the $p^{th}$ parameter of $\theta_{f-q}$, and ${(\theta_{f,Burn-in})}^*_p$ is its corresponding optimal value obtained during the burn-in stage.

\section{Related Work and Experiments}
To demonstrate the effectiveness of our proposed method, we now evaluate it against a wide range of alternatives, to address the extremely challenging, yet under-studied scenario of object detection at night-time. For this purpose, we use the recently proposed, large-scale BDD100K dataset \cite{BDD100k_CVPR20} with its original class labels, and filter out the images that belong to day/ night-time. This results in the following data splits (along with respective number of images within bracket): i) Training daytime (36728), ii) Training night-time (27971), iii) Validation daytime (5258), and iv) Validation night-time (3929). There are 9 categories (person, rider, car, bus, truck, bike, motor, traffic light, traffic sign) that are present in the images and are suitable for training (the \textit{train} category is omitted due to very few instances). As for our base object detector, and without loss of generality, we employ the single-stage YOLOF \cite{yolof_cvpr21} method.

\textbf{Justification of the choice of our dataset: } The BDD100K dataset actually presents diverse conditions present in real-world nighttime images. The best part is the presence of a large number of 2D boxes. Also, the images are a good mix of both high- vs low-resolution, and near vs far away aspects of objects. At the same time, while the Average Precision (AP) performance metric of state-of-the-art object detectors on standard benchmarks like COCO \cite{COCO_eccv14} is high (37-55 using various backbones), the same is very poor in BDD100K (around 17-24 when trained following our evaluation protocol. We noted that some methods reported high performance by considering only one or two large object categories, like cars, but, on the other hand we report results across all 9 object categories, even for the smaller categories, like rider, person, etc, which lowers the overall average). BDD100K captures real-world day-night corner cases, is large enough, and thus provides significant room for exploration.

On the contrary, similar datasets have their problems of their own, to find suitability in our evaluation. The KAIST dataset \cite{KAIST_dataset_CVPR15} consists of color-thermal pairs, but has annotations only for certain categories (person, people, cyclist). Nightowls dataset \cite{Nightowls} has annotated images at night-time, but only for pedestrians. For a day-night coupling, it also needs a separate daytime pedestrian dataset like Caltech Pedestrian \cite{Caltech_Pedestrian_TPAMI12}. In any case, a single category does not make up a decent evaluation protocol. The Waymo open \cite{Waymo_open_CVPR20} and nuScenes \cite{nuscenes_cvpr20} datasets provide 3D labels, while we focus on the more commonly used 2D boxes based methods. The Dark Zurich dataset \cite{semseg_nighttime_curriculum_iccv19} only provides GPS labels for day-twilight-night correspondence. The ACDC dataset \cite{ACDC_dataset_21} provides semantic segmentation labels, while we focus on object detection. The Alderdey dataset \cite{Alderdey_dataset} has day-night correspondences, but no annotations.

\textbf{Justification of the choice of our object detector: } There are a countless number of object detector backbones in literature, and studying different choices of backbones is neither the focus nor under the scope of our paper. Compared to the widely studied 2-stage Faster RCNN method, we preferred single-stage models, due to their speed, which is more practical for real-world situations, such as autonomous driving. Reasons for choosing YOLOF (over recent, state-of-the-art alternatives such as RetinaNet \cite{focalloss_iccv17}/ YOLOv4 \cite{scaled_yolov4_cvpr21}/ transformer based DETR \cite{DETR_ECCV20}): It outperforms others in similar settings \cite{yolof_cvpr21}, can be trained in much lesser epochs, requires lesser GFLOPs, and also has higher FPS. It is also simpler than YOLOR \cite{wang2021you}, and is built in Detectron2. While there are multiple tools available to implement object detectors, we make use of Detectron2, that i) supports a number of computer vision research projects, ii) provides a much cleaner way to incorporate new capabilities, and iii) facilitates much faster training.

\textbf{Generative approaches for synthetic images and Image Enhancement as alternatives? } Generative Adversarial Networks (GAN) based unpaired Image-to-Image (I2I) translation techniques such as UNIT \cite{UNIT_NeurIPS17}, MUNIT \cite{MUNIT_ECCV18}, ToDayGAN \cite{todaygan_icra19}, ToNightGAN\cite{tonightgan_ijcnn19}, CoMoGAN \cite{comogan_cvpr21} are quite popular to convert daytime images to other domains (eg, night-time) and use the corresponding bounding boxes from labeled daytime images for detector training with the synthetic images. To further strengthen the generative translation, DUNIT \cite{DUNIT_CVPR20}, in addition to global features, considers instance level features, whereas ForkGAN \cite{forkgan_eccv20} and AU-GAN \cite{AUGAN_BMVC21} consider sophisticated architectural changes for adversarial training. Techniques like AugGAN \cite{AugGAN_ECCV18} and Multimodal-AugGAN/ MultiAugGAN \cite{Multimodal_AugGAN_AAAI20} require additional semantic segmentation annotation, apart from bounding box annotations (though costly in real-world settings).
\begin{figure}[!ht]
\centering
	\includegraphics[width=\linewidth]{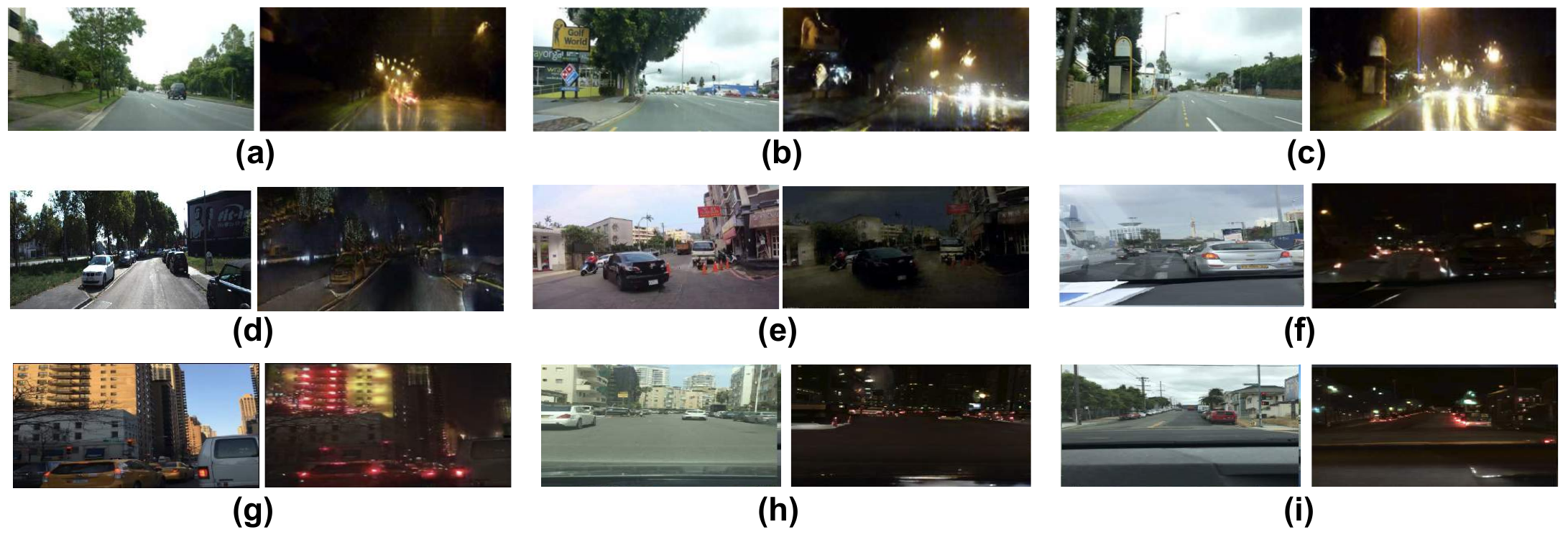}
    \caption{Common issues with GAN based methods for day-to-night image translation (left: day, right: night): a)-c) Disappearance of vehicles, with unnatural illumination generation throughout image, d)-e) appearance of dark patches, f) disappearance of small, far objects (eg, bike rider) due to too much blackening of image, g)-i) appearance of lights in abnormal locations (eg, throughout the balcony of the building in g) ). NOTE: Fig is best viewed in color monitor by zooming-in.}
    \label{gan_problems}
\vspace{-0.7cm}
\end{figure}

While GAN based image translation produces decent-looking night-images, they often lead to creation of unrealistic/ semantically irrelevant artifacts (not at all present in the original day image), see Fig \ref{gan_problems}. For eg, as shown in Fig \ref{gan_problems}-a), a car present in the daytime image is completely vanished in the translated night-image, and is replaced by extreme illumination. Now, training a detector with a bounding box on that location would only misguide the model. At the same time, as an alternative to GAN based translation, while it may also be tempting to simply apply recent, state-of-the-art image illumination/ enhancement techniques (such as EnlightenGAN \cite{enlightengan_tip21}, Zero-DCE \cite{zerodce_tpami21}, KinD \cite{KinD_ACMMM19}) on inference images from night, we later show that it seldom works well in practice.

\textbf{Training Configuration:} To demonstrate the effectiveness of our proposed method, we compare it against a number of recent, related, State-Of-The-Art (SOTA) \textbf{Image Enhancement} (KinD \cite{KinD_ACMMM19}, ZeroDCE \cite{zerodce_tpami21}, EnlightenGAN \cite{enlightengan_tip21}), GAN based \textbf{translation} (ToDayGAN \cite{todaygan_icra19}, CoMoGAN \cite{comogan_cvpr21}, AU-GAN \cite{AUGAN_BMVC21}, ForkGAN \cite{forkgan_eccv20}), GAN based \textbf{translation with instance feature extraction} (DUNIT \cite{DUNIT_CVPR20}), and GAN based \textbf{translation with auxiliary task based augmentation methods} (AugGAN \cite{AugGAN_ECCV18}, MultiAugGAN \cite{Multimodal_AugGAN_AAAI20}).

For each of these baseline methods, we make use of their corresponding best hyperparameters suggested by the original papers, and generate day-time to night synthetic images, for the GAN methods. These synthetic night images are then used to train our base YOLOF object detector, with the corresponding bounding boxes from day-time. The trained models are then used to infer upon the test nighttime images. For the image enhancement methods, we use the test nighttime images for image enhancement, before inferring upon them.

To make sure that all methods (including ours) have been trained in an \textbf{uniform protocol}, for the object detector, we make use of the YOLOF\_R\_50\_C5\_1x \cite{yolof_cvpr21} model \footnote{\url{https://github.com/chensnathan/YOLOF}}, with the PyTorch based Detectron2 tool, and fix a batch size of 16, and base learning rate of 0.025 (and initial warmup iterations of 1500) for all the experiments. This set of hyperparameters is feasible for single GPU training (on a Tesla V100-16GB). We do not make use of any learning rate decay. Rest of the hyperparameters are used as default from the Detectron2 tool. Please note that we collectively use the backbone and dilated encoder of the YOLOF model as our Encoder.
\begin{table*}[!t]
\centering
\caption{Performances of SOTA Object Detectors drop when a model trained in Domain A (eg, Day) is used for inference on images from Domain B with a huge domain mismatch (eg, Night). Day2Day: Train on Day, Test on Day; Day2Night: Train on Day, Test on Night; Night2Day: Train on Night, Test on Day; Night2Night: Train on Night, Test on Night. Results are obtained using the recent large-scale BDD100K dataset. We are interested in performing better than Day2Night.}
\label{motivation_DA}
\resizebox{0.9\linewidth}{!}{%
\begin{tabular}{@{}cccccccccccccccc@{}}
\toprule
\multicolumn{1}{c|}{Setting}                                   & AP                            & AP50                          & AP75                          & APS                          & APM                           & \multicolumn{1}{c|}{APL}                           & person                        & rider                        & car                           & bus                           & truck                         & bike                          & motor                         & traffic light                 & traffic sign                  \\ \midrule
\multicolumn{16}{c}{\cellcolor[HTML]{EFEFEF}YOLOF R50\_C5\_1x (1-Stage Object Detector)}                                                                                                                                                                                                                                                                                                                                                                                                                                                                                          \\ \midrule
\multicolumn{1}{c|}{Day2Day}                                   & 24.32                         & 46.88                         & 21.88                         & 6.08                         & 31.62                         & \multicolumn{1}{c|}{52.61}                         & 20.82                         & 16.87                        & 38.95                         & 37.79                         & 36.00                         & 15.86                         & 15.18                         & 13.98                         & 23.47                         \\
\multicolumn{1}{c|}{{\color[HTML]{FE0000} \textbf{Day2Night}}} & \cellcolor[HTML]{FFCCC9}17.59 & \cellcolor[HTML]{FFCCC9}35.66 & \cellcolor[HTML]{FFCCC9}15.32 & \cellcolor[HTML]{FFCCC9}4.50 & \cellcolor[HTML]{FFCCC9}20.44 & \multicolumn{1}{c|}{\cellcolor[HTML]{FFCCC9}36.15} & \cellcolor[HTML]{FFCCC9}16.03 & \cellcolor[HTML]{FFCCC9}8.09 & \cellcolor[HTML]{FFCCC9}28.51 & \cellcolor[HTML]{FFCCC9}27.21 & \cellcolor[HTML]{FFCCC9}26.51 & \cellcolor[HTML]{FFCCC9}13.85 & \cellcolor[HTML]{FFCCC9}10.05 & \cellcolor[HTML]{FFCCC9}6.13  & \cellcolor[HTML]{FFCCC9}21.89 \\ \midrule
\multicolumn{1}{c|}{Night2Night}                               & 19.75                         & 41.39                         & 16.43                         & 5.33                         & 22.14                         & \multicolumn{1}{c|}{41.33}                         & 15.75                         & 10.58                        & 33.29                         & 30.30                         & 30.30                         & 12.44                         & 12.24                         & 9.15                          & 23.72                         \\
\multicolumn{1}{c|}{Night2Day}                                 & 17.45                         & 35.61                         & 14.99                         & 3.44                         & 22.73                         & \multicolumn{1}{c|}{39.97}                         & 15.35                         & 12.19                        & 34.02                         & 25.19                         & 24.36                         & 10.66                         & 8.24                          & 8.90                          & 18.10                         \\ \bottomrule
\end{tabular}%
}
\vspace{-0.7cm}
\end{table*}

\textbf{Performance bounds:} To motivate the reader of the challenge in object detection at nighttime, we use Table \ref{motivation_DA}), to highlight the results of our initial experiment, where we showcase the results of 4 evaluations: Day2Day, Day2Night, Night2Night and Night2Day. The first two indicate that the YOLOF model was trained for 20k iterations on the entire labeled \textit{Training daytime} images, and evaluated on Validation daytime and Validation night-time splits respectively. The latter two indicate training for the same number of iterations using the entire set of labeled \textit{Training night-time} images, and evaluated on Validation night-time and Validation daytime splits respectively.

We could clearly see that using commonly used AP metrics, compared to Day2Day results, Day2Night results are poorer, and compared to Night2Night results, Night2Day results are poorer. This motivates the reader that a SOTA detector trained on labeled daytime images may not perform well on night-time inference images. Here, Day2Night and Night2Night represent the lower and upper bounds for the day to night evaluation scenario. For our method, we first train the detector only with labeled Training day images using the configuration mentioned earlier for Table \ref{motivation_DA}. The, we load the checkpoint and now also use synthetic images, and see how much better we can do.

\textbf{Fourier Transformation for synthetic images: } Due to the limitations of the GAN based approaches to generate synthetic images as seen earlier, we use frequency domain information via Fourier Transformation (FT) to obtain $x_i^{d\rightarrow s}$. Compared to generative image translation, FT does not require any training, and also produces better semantics preserving images. To obtain $x_i^{d\rightarrow s}$ using $x_i^d \in \mathcal{D}_d$, we need a random image $x_i^n \in \mathcal{D}_n$ (in our case, we use a \textit{seed night image}), to swap the \textit{style} (provided by amplitude) of $x_i^d$ with that of $x_i^n$. Let, $\mathcal{F}_a$ and $\mathcal{F}_p$ respectively denote the amplitude and phase components of the Fourier transformation $\mathcal{F}$ of an image $x_i$. Then, we can obtain $x_i^{d\rightarrow s}$ as \cite{fda_semseg_CVPR20}:
\begin{equation}
    \label{fourier_swap}
    x_i^{d\rightarrow s}=\mathcal{F}^{-1}([ M_\beta \circ \mathcal{F}_a (x_i^n) + (1-M_\beta) \circ \mathcal{F}_a (x_i^d) , \mathcal{F}_p(x_i^d) ]).
\end{equation}
Here, $\mathcal{F}^{-1}$ denotes the inverse Fourier transformation for mapping back the signals to the image space. $M_\beta$ is a mask with values zero except at a square (of size $\beta$) located at the center (0,0) of the amplitude signals. Here, $\beta \in (0,1)$.
\begin{figure}[!t]
  \centering
  \includegraphics[width=0.6\columnwidth]{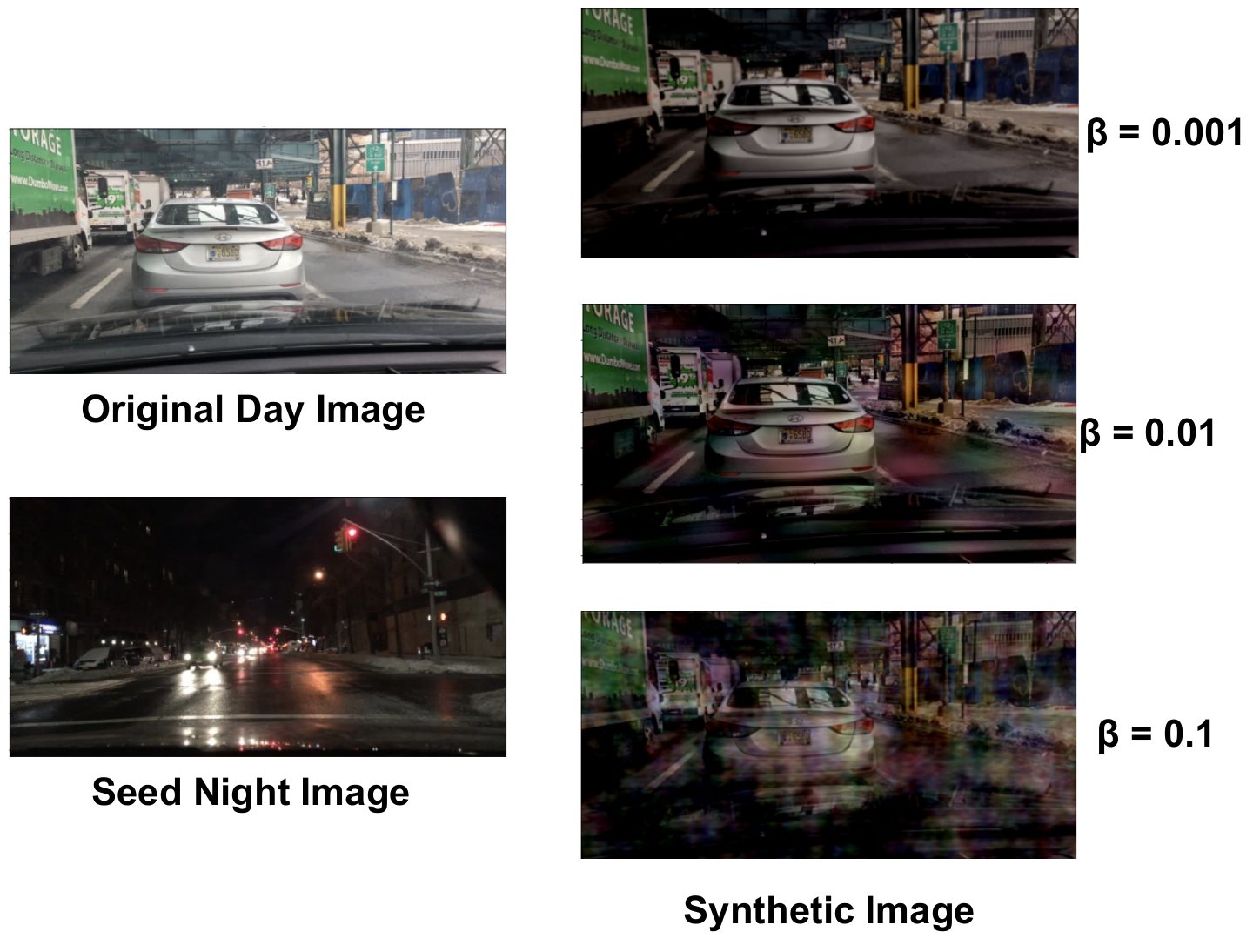}
   \caption{Illustration of Synthetic Image generation with Fourier Transformation (for various $\beta$ values).}
   \label{FT_quali_beta}
\vspace{-1.4cm}
\end{figure}

In Figure \ref{FT_quali_beta}, we showcase the results of synthetic image generation for a given original day image and a seed night image, for various $\beta$ values. The benefit of FT based synthesis is two-fold: i) \textbf{Computational}: While GAN based translation requires expensive training with sophisticated architectures, Fourier synthesis does not require any training, and can be done in merely a second on a CPU, ii) While GAN based translation often leads to creation of unrealistic/ semantically irrelevant artifacts (not at all present in the original day image), as shown in Fig \ref{FT_quali_beta}, \textbf{Fourier transformation preserves the semantics/ contours better}.

The quality of $x_i^{d\rightarrow s}$ depends on \textbf{three factors}: i) The anchor daytime image $x_i^d \in \mathcal{D}_d$, ii) The illumination present in the \textit{seed night-time} image $x_i^n \in \mathcal{D}_n$, and iii) The hyperparameter $\beta$ in (\ref{fourier_swap}). As a general trend, we observed that a lesser illuminated \textit{seed night-time} image and a higher $\beta$ value is more likely to produce a darker output image $x_i^{d\rightarrow s}$. However, setting $\beta$ very large also increases dark distortion in the output. We found $\beta=0.01$ to be mostly optimal in our case (with respect to lesser distortion). During translated image generation for training, we randomly use $\beta=0.01$ and $\beta=0.05$ (to account for darker images).

\textbf{Why training with synthetic images helps ?} The produced $x_i^{d\rightarrow s}$ (within a reasonable $\beta$) gives an impression of different training domains corresponding to different times/illumination of the day (depending on the darkness). This eventually improves the robustness of the model trained with Contrastive Learning, as we are teaching the model to focus on objects irrespective of the time of the day, with the hope that it will perform well on night images. This is analogous to \textit{Domain Generalization} in the classification literature \cite{domainbed_iclr2021} where a model is trained using labeled data from a number of training domains $\mathcal{L}_1,\cdots,\mathcal{L}_{tr}$. The idea is that, by learning statistical invariances among the domains, it will perform well on an unseen inference domain $\mathcal{L}_{tr+1}$. This is practically beneficial, because physically collecting labeled images from different times of the day is significantly tedious and expensive.

\subsection{Ablation Studies, and Detailed Hyperparameter Sensitivity Analysis of our method}
As part of our next experiments, we try to shed lights on each and every individual component of our method, with a focus on ablation studies, and a detailed hyperparameter sensitivity analysis.

\textbf{Generative image translation vs Fourier transformation for synthetic image generation: }
To study the broad alternatives for synthetic image generation, we first employ two GAN based translation methods like ToDayGAN \cite{todaygan_icra19} and CoMoGAN \cite{comogan_cvpr21} to generate synthetic images, which are then used to fine-tune the object detection model trained with the labeled daytime images. We also naively perform Fourier transformation based image synthesis, and then use those synthesized images to directly fine-tune the detector (instead of using our proposed Contrastive Learning module), referred to as Fourier Fine-Tune (Fourier FT). The results are presented in Table \ref{ablations_ganvsFT}. We could see that a worse performing generative translation involving sophisticated network training is no better than a simple Fourier transformation based image synthesis, which requires no training at all. Thus, for further experiments in our method, we employ Fourier transformation to generate the synthetic images for Contrastive Learning.
\begin{table}[!htb]
\vspace{-0.7cm}
\centering
\caption{Comparison of Generative vs Fourier Transformation based image synthesis for detector fine-tuning.}
\label{ablations_ganvsFT}
\resizebox{0.5\columnwidth}{!}{%
\begin{tabular}{l|cccccc}
\hline
\multicolumn{1}{c|}{Method}     & AP                                                   & AP50                                                 & AP75                                                 & APS                                                 & APM                                                  & APL                                                  \\ \hline
ToDayGAN                        & \cellcolor[HTML]{FFFFFF}{\color[HTML]{24292F} 14.95} & \cellcolor[HTML]{FFFFFF}{\color[HTML]{24292F} 32.29} & \cellcolor[HTML]{FFFFFF}{\color[HTML]{24292F} 11.73} & \cellcolor[HTML]{FFFFFF}{\color[HTML]{24292F} 3.48} & \cellcolor[HTML]{FFFFFF}{\color[HTML]{24292F} 17.15} & \cellcolor[HTML]{FFFFFF}{\color[HTML]{24292F} 33.66} \\
CoMoGAN                         & \cellcolor[HTML]{FFFFFF}{\color[HTML]{24292F} 15.82} & \cellcolor[HTML]{FFFFFF}{\color[HTML]{24292F} 32.59} & \cellcolor[HTML]{FFFFFF}{\color[HTML]{24292F} 14.05} & \cellcolor[HTML]{FFFFFF}{\color[HTML]{24292F} 3.26} & \cellcolor[HTML]{FFFFFF}{\color[HTML]{24292F} 17.43} & \cellcolor[HTML]{FFFFFF}{\color[HTML]{24292F} 34.34} \\ \hline
\multicolumn{1}{c|}{Fourier FT} & 15.89                                                & 33.19                                                & 13.70                                                & 4.19                                                & 18.99                                                & 33.15                                                \\ \hline
\end{tabular}%
}
\end{table}

\begin{figure}[!thb]
\vspace{-1.4cm}
\centering
  \includegraphics[width=0.6\columnwidth]{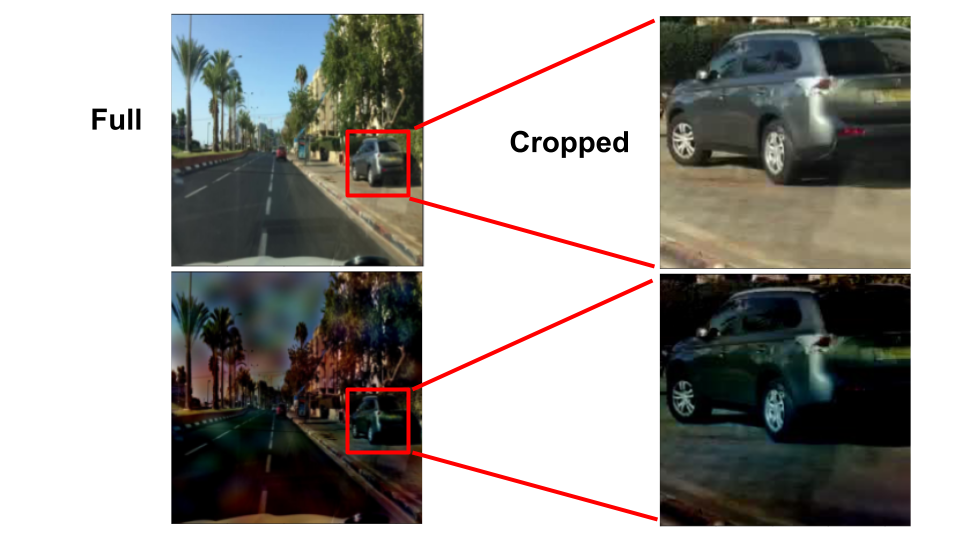}
  \captionof{figure}{Full and cropped images for Contrastive Learning.}
  \label{full_vs_crop}
\vspace{-0.6cm}
\end{figure}

\textbf{Benefit of Contrastive Learning, and whether to use cropping during Contrastive Learning ? } Now, we use the Fourier synthesized images for Contrastive Learning. But, we would now like to see if during this stage, we should be cropping the images. For this, we used \textit{only cropped images} (from same location in anchor-positive, see Fig \ref{full_vs_crop}) for data augmentation in training the Contrastive Learning module, but as shown in Table \ref{ablations_cropping} as \textit{Cropped}, it performs lower than \textit{Full}. This shows that merely using only random cropped images does not lead to a competitive AP (although for smaller objects, as shown by APS, it does lead to better performance). In practice, we would suggest randomly performing cropping with a very low probability, to account for smaller objects in a scene. At the same time, we could clearly see the benefit of our Contrastive Learning module, as the row corresponding to Full outperforms the Fourier FT method, which did not involve Contrastive Learning, but only fine-tuning of the detector.
\begin{table}[!t]
\centering
\caption{Ablation Studies showing role of cropping in Contrastive Learning.}
\label{ablations_cropping}
\resizebox{0.7\columnwidth}{!}{%
\begin{tabular}{@{}cccccc|cccccc@{}}
\toprule
\multicolumn{1}{c|}{Method}      & Crop & \begin{tabular}[c]{@{}c@{}}Decoder \\ Update\end{tabular} & Contrastive & Continual & \begin{tabular}[c]{@{}c@{}}2-Way \\ Update\end{tabular} & AP    & AP50  & AP75  & APS  & APM   & APL            \\ \midrule
\multicolumn{6}{c|}{\begin{tabular}[c]{@{}c@{}}Fourier FT (Train detector with Fourier generated images \\ and bounding boxes without Continual Contrastive Learning)\end{tabular}}                                                      & 15.89 & 33.19 & 13.70 & 4.19 & 18.99 & 33.15          \\ \midrule
\multicolumn{1}{c|}{Cropped}     & Crop only                                                 & No                                                        & \boldcheckmark         & No        & No                                                      & 15.61 & 32.76 & 13.10 & 4.54 & 19.47 & 30.76          \\
\multicolumn{1}{c|}{Full}        & Crop+Full                                                 & No                                                        & \boldcheckmark         & No        & No                                                      & 16.87 & 34.59 & 14.42 & 4.24 & 19.82 & 34.63          \\ \bottomrule
\end{tabular}%
}
\vspace{-0.8cm}
\end{table}

\textbf{Benefit of EWC based Continual Learning, and strategies to perform it: } We now make use of Continual Learning based penalty. As shown in Table \ref{ablations_all} as EWC, we get further improvement over \textit{Full} (with only Contrastive Learning), showing the benefit of Continual Learning. Despite the benefit of Continual Contrastive learning in Stage 2 \footnote{Update $\theta_f$ in EMA manner as: $\theta_f=\mu\theta_f+(1-\mu)\theta_{f-q}^*$, with $\mu=0.85$, where $\theta_{f-q}^*$ is feature extractor parameters from updated query in Stage-2.}, it is important to update the decoder in Stage-3. Otherwise, although the Encoder has been updated, the decoder would still carry weights from burn-in, and may not be consistent. Now, note that for Stage 1, we train the YOLOF model using the labeled Training daytime images, with the configuration discussed earlier. We load the model checkpoint obtained after Stage 1, for further use in Stages 2 and 3.
\begin{table}[!t]
\centering
\caption{Ablation Studies showing role of Continual Learning.}
\label{ablations_all}
\resizebox{0.7\columnwidth}{!}{%
\begin{tabular}{@{}cccccc|cccccc@{}}
\toprule
\multicolumn{1}{c|}{Method}      & Crop & \begin{tabular}[c]{@{}c@{}}Decoder \\ Update\end{tabular} & Contrastive & Continual & \begin{tabular}[c]{@{}c@{}}2-Way \\ Update\end{tabular} & AP    & AP50  & AP75  & APS  & APM   & APL            \\ \midrule
\multicolumn{1}{c|}{Full}        & Crop+Full                                                 & No                                                        & \boldcheckmark         & No        & No                                                      & 16.87 & 34.59 & 14.42 & 4.24 & 19.82 & 34.63          \\
\multicolumn{1}{c|}{EWC}         & Crop+Full                                                 & No                                                        & \boldcheckmark         & \boldcheckmark       & No                                                      & 17.43 & 35.44 & 15.11 & 4.19 & 20.29 & 36.00          \\
\multicolumn{1}{c|}{CL EWC}    & Crop+Full                                                 & \boldcheckmark                                                       & \boldcheckmark         & \boldcheckmark       & No                                                      & 18.38 & 37.33 & 15.69 & 4.34 & 20.20 & 39.78          \\
\multicolumn{1}{c|}{CL EWC 2w} & Crop+Full                                                 & \boldcheckmark                                                       & \boldcheckmark         & \boldcheckmark       & \boldcheckmark                                                     & \textbf{19.57} & \textbf{39.32} & \textbf{17.29} & \textbf{5.19} & \textbf{21.88} & \textbf{41.79} \\ \bottomrule
\end{tabular}%
}
\vspace{-0.8cm}
\end{table}

After the burn-in (Stage-1), the Stage-2 and Stage-3 interplay can be maintained in 2 different ways, resulting in two variants of our method: i) CL EWC, and ii) CL EWC 2 way (CL EWC 2 w). For Stage-2, using a sampled mini-batch (size 32), we backpropagate and update the query model, followed by EMA update of the key using (\ref{ema_FCL}). After that, for Stage-3, we can use the updated encoder of the query, to replace the counterparts in the object detector, and update only the Decoder (using mini-batch sampled for object detection as earlier, with labeled day images). Stage-2/-3 when alternated as mentioned, refers to CL EWC. Here, Contrastive Learning with EWC (CL EWC) denotes our proposed overall method.

Alternately, for Stage-3, we can use the updated encoder of the query, to replace the counterparts in the object detector, and update all of encoder and Decoder. Now, this updated encoder (denoted as $\theta_f^*$) can be used to further update the counterparts $\theta_{f-q}$ of the query in EMA manner as: $\theta_{f-q}=\mu\theta_{f-q}+(1-\mu)\theta_f^*$, with $\mu=0.99$, before updating the query again using Stage-2 of the next iteration. As observed, by using the Decoder update, CL EWC outperforms EWC (with Continual Contrastive learning in Stage 2, but without Stage 3). Also, by employing a 2-way update, CL EWC 2w outperforms CL EWC.

We also study the convergence behaviour of Stage-2 alone, by initializing with model from burn-in, and training just the contrastive loss. Figure \ref{losses_cl} displays the convergence behaviour of Contrastive Learning, yet again demonstrating its benefit over other alternatives such as adversarial learning \cite{dann_jmlr16} for aligning domains, when it comes to enjoying better convergence.

\subsection{Comparison with state-of-the-art: } In Table \ref{vs_SOTA_all}, we now report the final detection performance of our method against the SOTA methods discussed earlier, as our baselines. To obtain the results for our method reported in Table \ref{vs_SOTA_all}, Stage-2 of Contrastive learning was trained using batch size of 32, $\mu=0.99$ in (\ref{ema_FCL}), $\lambda_{ewc}=0.9$ in (\ref{overall_objective}), SGD optimizer with learning rate $5.5666945e-06$ (kept same as object detector, obtained using Detectron2 framework), weight decay of $1e-6$, momentum of $0.9$.
\begin{table*}[!htb]
\vspace{-0.9cm}
\centering
\caption{Comparison of our method against various SOTA translation and enhancement approaches. Metrics corresponding to the best method are shown in bold.}
\label{vs_SOTA_all}
\resizebox{\columnwidth}{!}{%
\begin{tabular}{@{}c|c|cccccc|ccccccccc@{}}
\toprule
Method                      & Nature                                                      & AP             & AP50                         & AP75           & APS           & APM            & APL            & person         & rider          & car            & bus            & truck          & bike           & motor          & traffic light & traffic sign   \\ \midrule
\rowcolor[HTML]{EFEFEF} 
Day2Night                   & Baseline                                                    & 17.59          & {\color[HTML]{000000} 35.66} & 15.32          & 4.50          & 20.44          & 36.15          & 16.03          & 8.09           & 28.51          & 27.21          & 26.51          & 13.85          & 10.05          & 6.13          & 21.89          \\ \midrule
KinD \cite{KinD_ACMMM19}             & Enhancement                                                 & 12.91          & 26.87                        & 10.66          & 3.35          & 15.34          & 26.32          & 12.24          & 6.48           & 21.48          & 23.29          & 18.46          & 11.26          & 4.12           & 3.29          & 15.57          \\
ZeroDCE \cite{zerodce_tpami21}          & Enhancement                                                 & 12.12          & 26.17                        & 9.89           & 3.52          & 15.40          & 23.85          & 11.19          & 6.11           & 19.14          & 19.61          & 18.10          & 9.87           & 2.58           & 5.04          & 17.40          \\
EnlightenGAN \cite{enlightengan_tip21}       & Enhancement                                                 & 12.50          & 25.89                        & 10.44          & 3.29          & 14.15          & 24.91          & 12.34          & 6.48           & 21.98          & 21.31          & 18.43          & 11.27          & 2.76           & 2.34          & 15.58          \\ \midrule
ToDayGAN \cite{todaygan_icra19}          & Translation                                                 & 14.95          & 32.29                        & 11.73          & 3.48          & 17.15          & 33.66          & 14.03          & 8.81           & 25.67          & 22.70          & 21.78          & 13.79          & 4.77           & 5.01          & 17.95          \\
CoMoGAN-infer \cite{comogan_cvpr21}     & Translation                                                 & 11.81          & 25.42                        & 9.49           & 2.14          & 12.98          & 27.25          & 13.34          & 5.74           & 20.37          & 17.14          & 16.20          & 10.10          & 5.31           & 2.26          & 15.82          \\
CoMoGAN  \cite{comogan_cvpr21}           & Translation                                                 & 15.82          & 32.59                        & 14.05          & 3.26          & 17.43          & 34.34          & 15.33          & 10.22          & 28.65          & 23.84          & 23.00          & 12.09          & 7.19           & 4.13          & 17.94          \\
AU-GAN \cite{AUGAN_BMVC21} & Translation &17.33	&35.73	&14.74	&3.71	&19.25	&39.48	&15.85	&9.11	&28.44	&25.95	&26.14	&13.55	&10.38	&6.05	&20.51 \\
DUNIT  \cite{DUNIT_CVPR20}           & Trans/Instance                                                 & 17.70	       &35.29	         &16.11     	&3.94	        &19.61	        &39.70	        &15.93	        &9.74	        &28.37	        &28.05	        &25.92	        &13.91	        &9.53	        &6.79	        &21.10          \\
AugGAN  \cite{AugGAN_ECCV18}           & Trans/Augment                                                &17.78	&36.45	&15.66	&4.26	&20.17	&36.82	&16.51	&8.60	&29.53	&27.38	&25.75	&16.76	&7.78	&6.50	&21.20          \\
MultiAugGAN  \cite{Multimodal_AugGAN_AAAI20}           & Trans/Augment    &18.55	&\textbf{39.81}	&14.89	&4.69	&20.66	&39.50	&13.66	&9.94	&\textbf{32.02}	&29.52	&27.34	&11.56	&10.13	&\textbf{8.90}	&\textbf{23.85}          \\
ForkGAN \cite{forkgan_eccv20}           & Translation                                                 & 18.24          & 36.63                        & 15.73          & 4.78          & 20.47          & 39.53          & 16.19          & 9.81           & 29.66          & 29.23          & 27.66          & 13.39          & 9.16           & 6.55          & 22.51          \\ \midrule
\rowcolor[HTML]{CBCEFB} 
\textbf{CL EWC 2w (Ours)} & \begin{tabular}[c]{@{}c@{}}Feature \\ Learning\end{tabular} & \textbf{19.57} & 39.32               & \textbf{17.29} & \textbf{5.19} & \textbf{21.88} & \textbf{41.79} & \textbf{17.72} & \textbf{10.52} & 30.48 & \textbf{30.59} & \textbf{27.77} & \textbf{17.17} & \textbf{11.08} & 7.36 & 23.47 \\ \bottomrule
\end{tabular}%
}
\vspace{-0.9cm}
\end{table*}

Also, as for the architectural configuration in Figure \ref{FCLUDA_OD}, we have an average pool plus flattening layer between the encoder and projector in both the query and key branches. Common to both, we first have a FC\_pro1(in=512,out=512) layer, followed by batchnorm and ReLU, follwed by FC\_pro2(in=512,out=128) layer. For the predictor, we have a FC\_pre1(in=128,out=512) layer, followed by batchnorm and ReLU, and followed by a FC\_pre2(in=512,out=128) layer. For data augmentation, we randomly used the whole image 75\% of the time, and performed random cropping the remaining 25\% (the same crop was applied to both day, and the synthetic image). Joint Stage-2/-3 training was performed for 14000 iterations. Note that here FC\_pro1, FC\_pro2, FC\_pre1 and FC\_pre2 are fully-connected layers with input size provided by ``in=" and output size provided by ``out=".

$\mu=0.99$ in (\ref{ema_FCL}) is fairly a standard value in EMA based updates in literature (eg, semi/self-supervised learning, etc). Similar role is played by $\lambda_{ewc}$ in (\ref{overall_objective}). Sensitivity analysis of $\lambda_{ewc}$: 1-2 pp drop is observed for values lesser than 0.65. But, for values within (0.75-0.95), the AP remains stable, with the highest observed for 0.9 (as reported). $\beta$ is not specific to object detection, but rather, the image synthesis. We recommend our suggested values ($\beta=0.01$ to $0.05$), where images generated have lesser artefacts. Night-time seed keeps picked randomly to enforce diversity.
\begin{figure}[!ht]
\vspace{-0.8cm}
\centering
  \includegraphics[width=0.7\columnwidth]{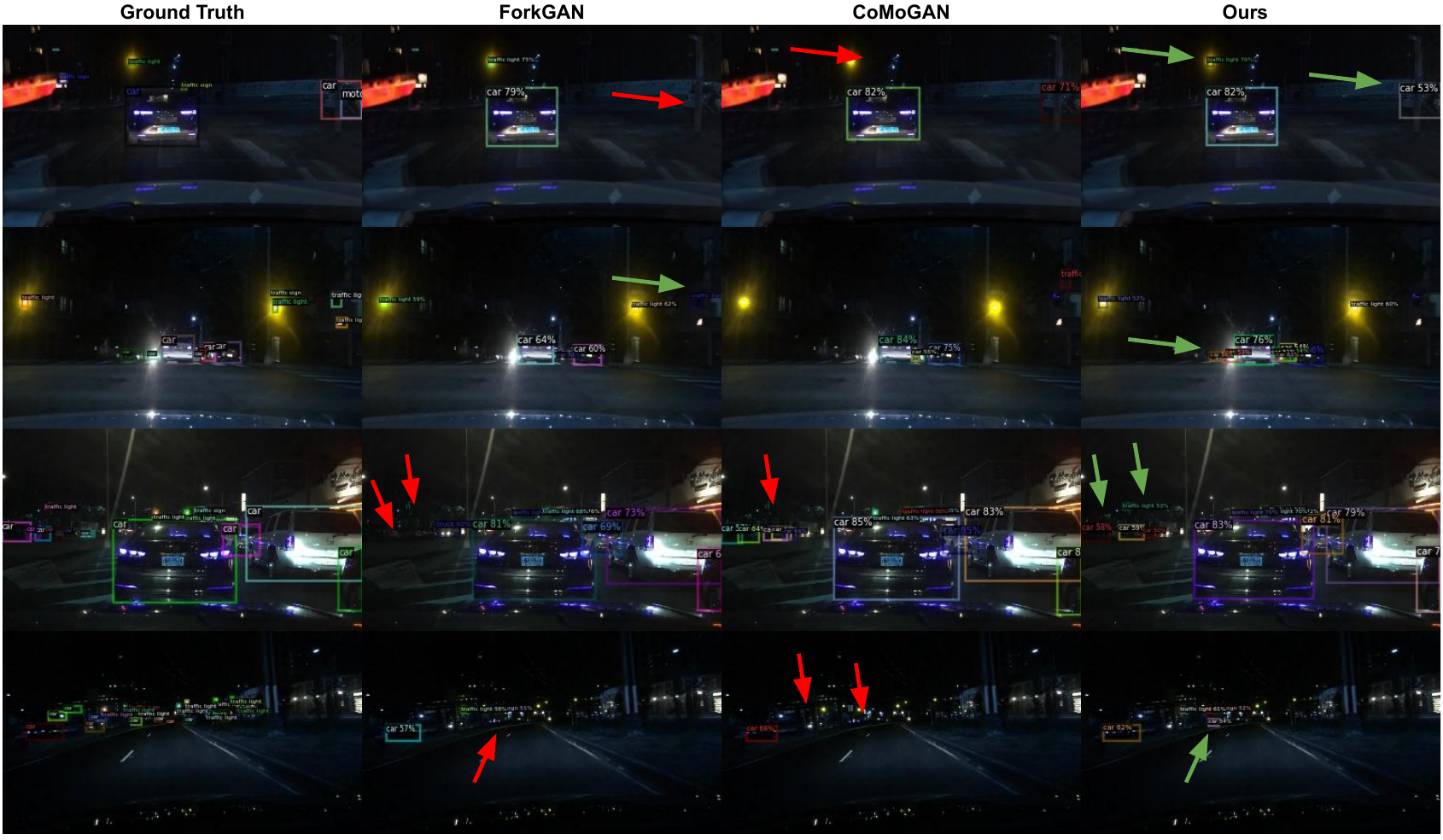}
	\includegraphics[width=0.7\columnwidth]{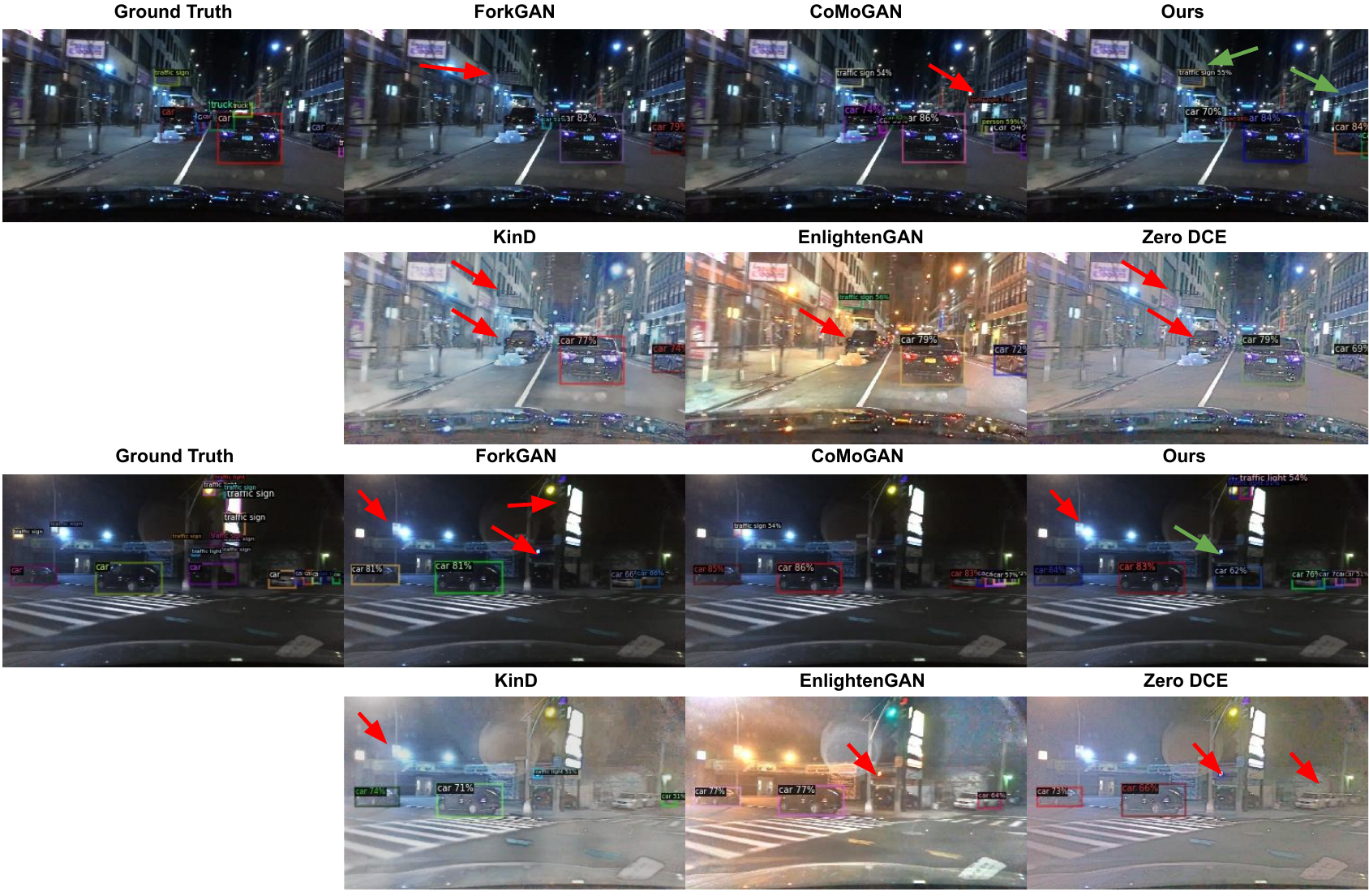}
    \caption{Qualitative comparison among various methods (best viewed: zoomed in color). Red arrows indicate incorrect predictions (false negative/positive). A green arrow for a method indicates that it did not make a mistake, compared to others.}
    \label{quali_vs_all}
\vspace{-0.9cm}
\end{figure}

As reported in Table \ref{vs_SOTA_all}, compared to the lower bound of this setting, i.e, Day2Night, image enhancement methods perform poor when only used to illuminate the night images before being inferred upon by the model trained with labeled daytime images. Also, ToDayGAN and CoMoGAN translation methods perform poor as well. As illustrated qualitatively in Fig \ref{gan_problems}, in certain cases GAN based methods may lead to patches/ artifacts, which explains the poor performance of ToDayGAN/ CoMoGAN. We also tried a variant of CoMoGAN (CoMoGAN-infer) to simply translate inference images to daytime, and use the day time trained model, without any training, and found it to be inferior than CoMoGAN with training.

With a sophisticated architecture and adversarial loss, added with 2-stage image translation (and refinement) during inference, ForkGAN performs as a SOTA. DUNIT though separately extracts object level features, but eventually fuses it with the global feature for image generation, which is where the robustness gets lost. However, it performs competitive as well. AugGAN and MultiAugGAN both make use of additional segmentation masks for training, which explains their competitiveness. However, our proposed CL EWC 2w, by virtue of its training tactics at the feature level, alleviates possible shortcomings during image translation, and outperforms the others, even without using any auxiliary information like MultiAugGAN, or sophisticated architecture like the ForkGAN.

In Figure \ref{quali_vs_all}, we showcase a few qualitative detection results among various methods compared. For the very shortcomings of the compared methods as discussed, while they make certain mistakes (red arrows), our method does well in cases where others failed (shown by green arrows).

\section{Conclusion}
We propose a novel, contrastive learning method to align the latent representations of a pair of real and synthetic images, to make the detector robust to the different domains. However, we found that merely contrasting the embeddings may lead to catastrophic forgetting of the information essential for object detection. Hence, we employ a continual learning based penalty, to alleviate the issue of forgetting, while contrasting the representations. We showcase that our proposed method outperforms a wide range of alternatives to address the extremely challenging, yet under-studied scenario of object detection at night-time, which is essential for reliable smart camera systems.

\section*{Acknowledgment}
I would like to thank Professor Robby from the National University of Singapore for the insightful conversations.

{\footnotesize
\bibliographystyle{splncs04}
\bibliography{CLEWC_OD_ECCV22_arxiv}
}

\end{document}